\documentclass[10pt,twocolumn,letterpaper]{article}

\usepackage{cvpr}              

\usepackage{graphicx}
\usepackage{amsmath}
\usepackage{amssymb}
\usepackage{booktabs}
\usepackage{tabularx}
\usepackage{pifont}

\usepackage{colortbl}
\usepackage{multirow}

\usepackage{wrapfig}
\usepackage{makecell}
\usepackage[accsupp]{axessibility}  

\definecolor{bkr}{cmyk}{0,0,0,0.15}

\newcommand{\argmin}{\mathop{\mathrm{argmin}}\limits}

\usepackage[dvipsnames]{xcolor}

\usepackage[pagebackref=true,breaklinks=true,letterpaper=true,colorlinks,citecolor=ForestGreen, bookmarks=false]{hyperref}

\usepackage[capitalize]{cleveref}
\crefname{section}{Sec.}{Secs.}
\Crefname{section}{Section}{Sections}
\Crefname{table}{Table}{Tables}
\crefname{table}{Tab.}{Tabs.}

\newcommand\blfootnote[1]{%
  \begingroup
  \renewcommand\thefootnote{}\footnote{#1}%
  \addtocounter{footnote}{-1}%
  \endgroup
}

\newcommand{\Raise}[1]{\small\textcolor{ForestGreen}{\xspace{\bf $\uparrow$#1}}}
\newcommand{\Drop}[1]{\small\textcolor{Bittersweet}{\xspace{\bf $\downarrow$#1}}}

\def\cvprPaperID{6096} 
\def\confName{CVPR}
\def\confYear{2023}
\title{Detection Hub: Unifying Object Detection Datasets via Query Adaptation on Language Embedding}

\author{
  Lingchen Meng$^{1,2}$~\hspace{10pt}
  Xiyang Dai$^{3}$~\hspace{10pt}
  Yinpeng Chen$^{3}$~\hspace{10pt}
  Pengchuan Zhang$^{3}$~\hspace{10pt}
  Dongdong Chen$^{3}$~\hspace{10pt} \\
  Mengchen Liu$^{3}$~\hspace{10pt}
  Jianfeng Wang$^{3}$~\hspace{10pt}
  Zuxuan Wu$^{1,2\dagger}$~\hspace{10pt}
  Lu Yuan$^{3}$~\hspace{10pt}
  Yu-Gang Jiang$^{1,2}$  \vspace{0.1in}\\ 
$^{1}$Shanghai Key Lab of Intell. Info. Processing, School of CS, Fudan University \\
$^{2}$Shanghai Collaborative Innovation Center of Intelligent Visual Computing \\
$^{3}$Microsoft}

\begin{document}

\maketitle

\begin{abstract}
\blfootnote{$^{\dagger}$ Corresponding author.}

Combining multiple datasets enables performance boost on many computer vision tasks. But similar trend has not been witnessed in object detection when combining multiple datasets due to two inconsistencies among detection datasets: taxonomy difference and domain gap. In this paper, we address these challenges by a new design (named Detection Hub) that is \textit{dataset-aware} and \textit{category-aligned}. It not only mitigates the dataset inconsistency but also provides coherent guidance for the detector to learn across multiple datasets. In particular, the dataset-aware design is achieved by learning a dataset embedding that is used to adapt object queries as well as convolutional kernels in detection heads. The categories across datasets are semantically aligned into a unified space by replacing one-hot category representations with word embedding and leveraging the semantic coherence of language embedding. Detection Hub fulfills the benefits of large data on object detection. Experiments demonstrate that joint training on multiple datasets achieves significant performance gains over training on each dataset alone. Detection Hub further achieves SoTA performance on UODB benchmark with wide variety of datasets. 

\end{abstract}

\section{Introduction}

\begin{figure}[t]
	\begin{center}
		\includegraphics[width=1.0\linewidth]{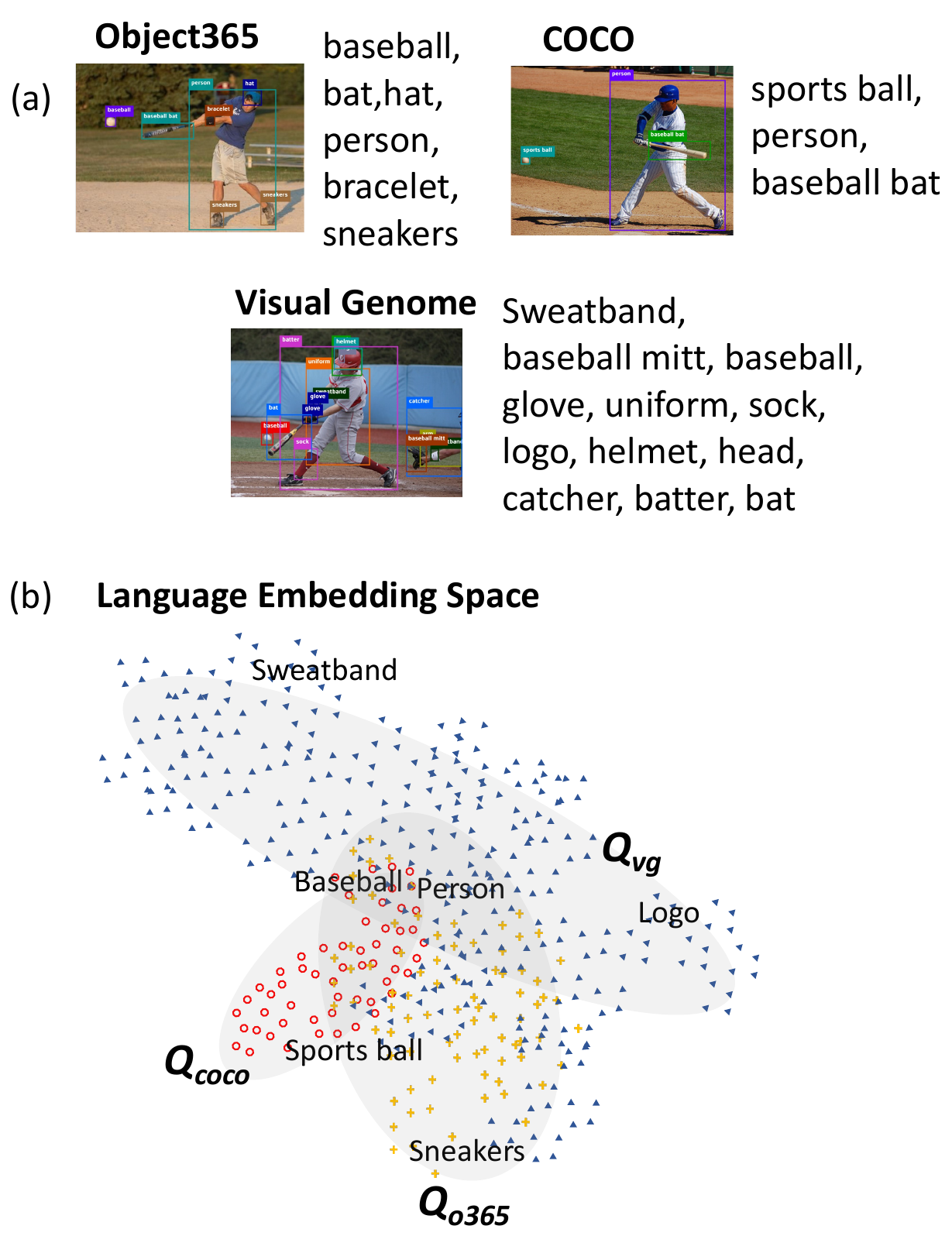}
	\end{center}
	    \vspace{-5mm}

	\caption{An illustration of the challenges of combing multiple datasets. (a) There are significant domain gaps between different datasets: the taxonomy are different and the annotations are inconsistent; (b) Even under a unified language embedding, the distribution of categories among datasets are significant different and requires special handling}
	\label{fig:teaser}
    \vspace{-5mm}
\end{figure}

Recent computer vision development has demonstrated significant benefits of leveraging large-scale data for computer vision tasks, such as image retrieval~\cite{clip}, image recognition~\cite{vinvl} and video recognition~\cite{bevt}. However, limited effort has been explored for the object detection task due to the lack of a unified large-scale data. A naive attempt is to combine all annotated data from different sources. However, due to the diversity of objects and the cost of annotating bounding boxes in images, traditional object detection datasets are collected in a domain-specific way, where a limited number of interested categories are regarded as foregrounds and the rest of objects as backgrounds. This results in a non-trivial domain shift between different datasets and limits detectors to be trained and tested on a single dataset in order to achieve the best performance on such dataset. 

In this paper, we attempt to answer the question: \emph{``How can we unify multiple object detection datasets training with a single general object detector?''}. Towards this goal, we observe two challenges: taxonomy difference and annotation inconsistency, both of which introduce the domain gap issue, shown in Figure \ref{fig:teaser} (a). More specifically, for the taxonomy difference issue, the semantic names of similar concepts in different datasets may be very different. And for the annotation inconsistency issue, given similar images, foreground objects in one dataset may be labeled as background in another dataset. The existence of these two challenges may be the key underlying reason why most current detectors only focus on a specific training set, rather than deriving a universal object detector for multiple datasets.

Compared to traditional methods that regard semantic categories as class indices, converting them into language embedding can naturally unify the taxonomy and eases the challenge in recent works \cite{glip,clip}. However, it won't solve the problem, as shown in Figure \ref{fig:teaser} (b), the distribution of categories among datasets are significant different, which contributes most to the domain gap. 

Inspired by recent success of visual-language models~\cite{glip,clip}, we propose a simple yet effective method \emph{``Detection Hub"} that can  enjoy the synergy of multiple datasets. It builds upon current end-to-end detectors which use learnable object queries to produce final detection results~\cite{sparse}. To overcome the aforementioned problems, we have two key ideas accordingly: 1) map the semantic category names of different datasets into a category-aligned embedding, and 2) more importantly, use the embedding to dynamically adapt object queries so that each dataset has its own specific query set. We further change the one-hot based classification branch into vision-language alignment, which can well align categories of different datasets. Accordingly, we adopt a region-to-word alignment loss instead of classical cross entropy loss, which makes our \emph{``Detection Hub"} not limited by a fixed category vocabulary.

Our method leverages the linguistic property of pretrained language encoders (such that categories with similar semantic meanings across datasets will be automatically embedded together without using an expert-designed label mappers). In addition, categories specific to different datasets will be preserved by dedicated embedding. Moreover, as each dataset has its own adapted object query set to generate detection results, the detector can learn how to adapt its behavior to each dataset based on its specific query set. Therefore, the potential competition or disturbance caused by the annotation inconsistency among different datasets can be avoided.

To demonstrate the effectiveness of our method, we train our \emph{``Detection Hub"} on three standard object detection datasets jointly: COCO~\cite{coco}, Object365~\cite{o365} and Visual-Genome~\cite{vg}. These large-scale datasets have different properties of taxonomy, vocabulary size and annotation quality. Detecton Hub achieves 45.3, 23.2 and 5.7 AP on each dataset, with significant performance gain of \textbf{+2.3, +1.0, +0.9} compared with each independently model. To further verify the effectiveness on datasets with larger variance, we conduct experiments on UODB~\cite{wang2019towards}, a combination of 11 extremely varied datasets. Detection Hub achieves an average score of 71 and outperforms the previous SoTA UniversalDA~\cite{wang2019towards} by a large margin of \textbf{6.8} point.

\section{Related Work}

\noindent\textbf{Object Detection.} Object detection~\cite{rcnn, fastrcnn, focal, fcos, fasterrcnn, rfcn, noisy_anchor} has been studied for a long time and become a predominant direction. Beginning with the initial success of R-CNN~\cite{rcnn}, a two-stage paradigm is formulated by combining a region proposal detector and a region-wise CNN classifier. Fast R-CNN~\cite{fastrcnn} formalizes the idea of region-wise feature extraction using ROI pooling to extract fixed-size region features and then predicts their classes and boxes with fully connected layers. Faster R-CNN~\cite{fasterrcnn} further introduces a Region Proposal Network (RPN) to generate region proposals and merges the two-stage into a single network by sharing their convolutional features. Since then, this two-stage architecture has become a leading object detection framework. Further works continue to improve this framework by introducing more stages~\cite{cascade, saccadenet}. Recently, Vision Transformers~\cite{vit} and end-to-end object detection methods have attracted such attention. DETR~\cite{detr} first turns the object detection problem into a query-based set prediction problem and formulates object detection as an end-to-end framework. Many follow-ups~\cite{adavit, mal, deform-detr, conditional-detr} further developed these methods into other tasks like acceleration and pseudo labeling. Beside, multiple recent methods were proposed to improve the problems in different directions. Deformable DETR~\cite{deform-detr} first introduces multi-scale deformable attention to replace transformers in DETR and reduces the key sampling points around the reference point to significantly accelerate the training speed. Dynamic DETR~\cite{dydetr} further models dynamic attention among multiple dimensions with both a convolution-based encoder and a decoder to further reduce learning difficulty and improves the performance. Sparse R-CNN~\cite{sparse} proposes a sparse set of learnable proposal boxes and a dynamic head to perform end-to-end detection upon the R-CNN framework. Our proposed method is also a query-based object detector. Unlike those DETR query designs, our linguistic adapted query is irrelevant to position and plays a role like a semantic-guided filter.

\vspace{0.05in}
\noindent\textbf{Multiple Datasets Training.} There are a few early attempts on training with multiple datasets to recognize a wide variety of objects. YOLO9000~\cite{yolo} first proposed a method to jointly train on object detection and image classification, which expanded the detection vocabulary via classification labels. Later,~\cite{det11k} expanded such an approach to two large scale datasets, ImageNet and OpenImages with only a small fraction of fully annotated classes.~\cite{wang2019towards} attempted to build a universal object detection system that is capable of working on
various image domains, from diverse datasets by introducing a series of adaptation layers based on the principles of squeeze and excitation.~\cite{unidet} further proposed to improve loss functions that integrated partial but correct annotations with complementary but noisy pseudo labels among different datasets. Most recently,~\cite{zhou2021simple} proposed to train a shared detector with dataset-specific heads and then learned to projected the outputs into a unified label space. Thanks to the advantages of leveraging language embedding, our method does not need to learn a joint label space to merge different datasets. This naturally addresses the scaling up of categories and joint of different datasets. 

\vspace{0.05in}
\noindent\textbf{Language Embedding.} Generative and efficient language representation is an attractive topic in the past few decades. As an integral role in the modern NLP system, word embedding can encode rich semantic in a structured way. Most widely used word embeddings methods, Word2Vec~\cite{word2vec}, GloVe~\cite{glove}, leverage the co-occurrence statistics in a large corpus. As a result, words with similar meanings are close in the embedding space, which can encode rich semantic information.
Furthermore, many follow-ups~\cite{elmo, gpt, bert}  extract context-sensitive features rather than static features. They perform sequence modeling with RNNs or Transformers on the top of traditional static word embeddings.
Most recently, learning visual representation with language embedding from language supervision is a promising trend due to rich semantic information and flexible transferable ability of natural language. As a milestone, CLIP ~\cite{clip} designs an image-text alignment pre-train task and performs contrastive learning on a large amount of image-text pairs. Motivated by CLIP, many following works~\cite{glip, regionclip, vild} try to improve the training strategy. GLIP~\cite{glip} reformulate the object detection as phrase grounding, which makes detection benefit from large grounding datasets. RegionCLIP~\cite{regionclip} and ViLD~\cite{vild} leverage the semantics space of CLIP for open-vocabulary object detection. Unlike previous methods attempting to learn a language embedding specifically, we freeze a pre-trained language embedding and design an adaption mechanism to dynamically adapt queries on categories based on the different distributions of datasets.

\section{Our Method}
\subsection{Revisiting End-to-end Object Detection}
End-to-end detectors~\cite{detr,dydetr,sparse} utilize object queries to encode the content and position statistics over the training dataset and drive the detector to predict desired objects localization without non-differential components (such as pre-defined anchors and non-maximum suppression). 

\begin{figure*}[!t] \centering
	\begin{center}
		\includegraphics[width=0.9\linewidth]{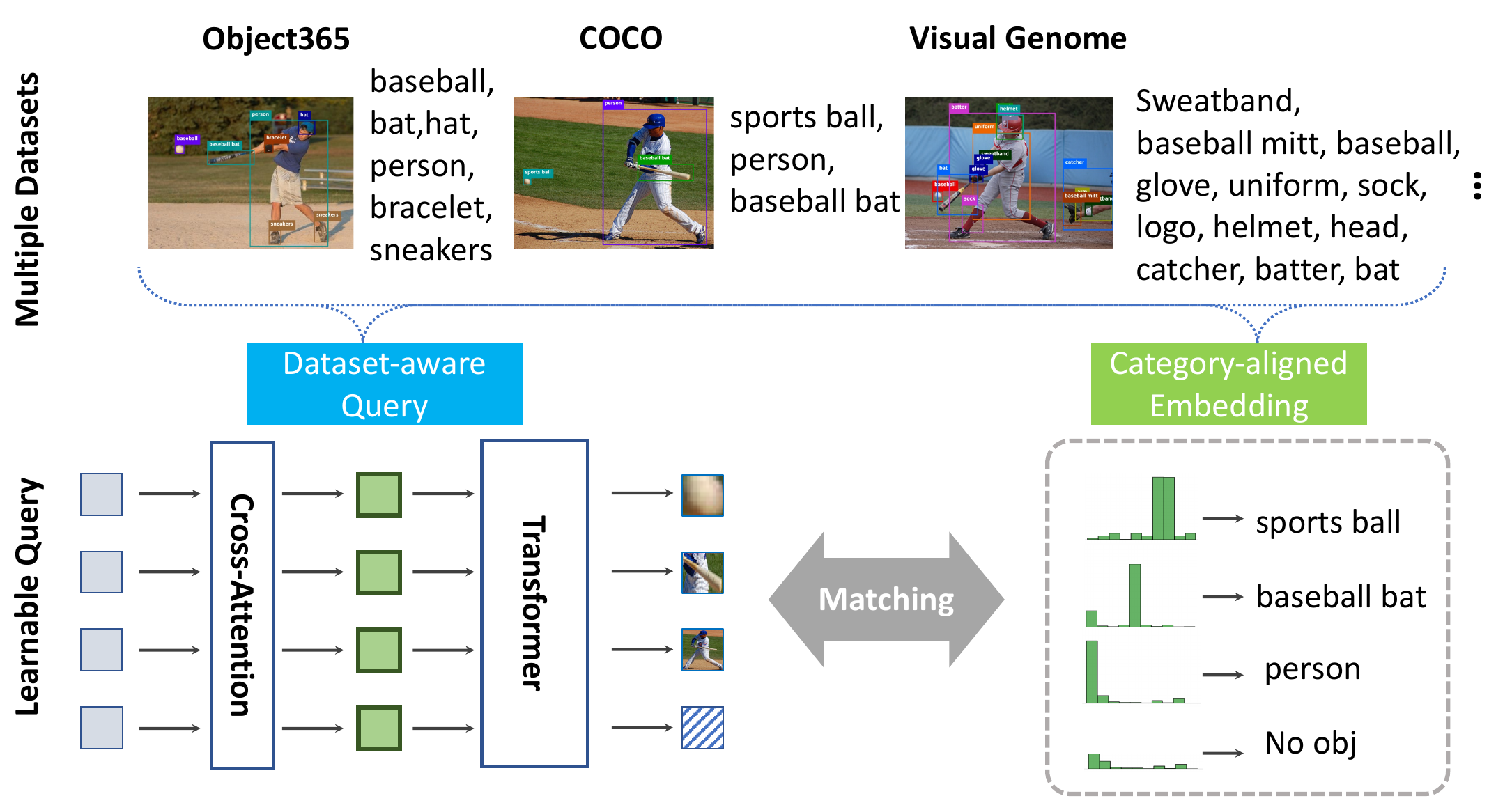}
	\end{center}
	\caption{An illustration of our Detection Hub design. Based on a category-aligned embedding, our dataset-aware queries are learned to dynamically adapt on category based on the different distributions of datasets. }
	\label{fig:arch}
    \vspace{-5mm}
\end{figure*}

Given a set of $N$ learnable queries $v \in \mathbb{R}^{N\times d}$, an end-to-end detector utilizes a  transformer $\mathcal{T}$ to generate $N$ corresponding predictions. Then the optimization process can be abstracted as solving a matching problem $L_{match}$ among images in the training dataset $D$:

\begin{equation}
    \sigma = \argmin_{\sigma \in Q_N} \sum^D_{i} L_{match}(y_i,\mathcal{T}(Q_N,i))
\end{equation}

Ideally, these $N$ learnable queries $Q_N$ can cover an arbitrary number of boxes in each image. However, considering $Q_N$ was optimized on the whole dataset, it will cause each object query to be in charge of multiple boxes and categories, meanwhile competing with each other to overfit to the training data \cite{anchor-detr, conditional-detr}. This limitation will be further amplified when learning across multiple datasets.

In this paper, we build upon Sparse-RCNN~\cite{sparse} as our default instantiation due to its intuitive explanation and training efficiency. Although achieving great performance on object detection, Sparse-RCNN has limitations on multi-dataset training: the learnable queries and features are dataset-dependant, which suffer from inconsistencies when the object statistics of datasets vary. Moreover, the classification branch is dataset-specific and hard to be jointly learned across different datasets. To enjoy the synergy and avoid such limitations, we propose category-aligned embedding and use it to adapt the object queries for each dataset to effectively formulate dataset-aware queries. In this way, our \emph{``Detection Hub"} can learn how to adapt its behavior for each dataset based on its own query set.

\subsection{Category-aligned Embedding}
Given two datasets $D_1$, $D_2$ with $n_1$, $n_2$ categories respectfully, we desire to have a embedding function $\mathcal{E}$ that can map them together in a common space. Traditional one-hot embedding cannot serve this purpose as it treats category solely as an index. Later works tried to manually or automatically learn a mapping function across datasets with limited success \cite{lambert2020mseg}.

It is natural to think about using category names directly, \emph{e.g.}, \cite{clip,unidet} use language models to generate unified embedding.  
In our method, we further leverage the semantic properties of the category set, through using a pretrained language model to encode the language embedding. Specifically, we concatenate the category names from a dataset $C$ together and then use a language model to obtain the language embedding $E$ of the dataset:
\begin{equation}
    \begin{split}
        &\text{CategorySet} = ``c_1, c_2, c_3, ..., c_n'' \\
        &E = \texttt{Embed}(\texttt{Tokenizer}(\text{CategorySet}))
    \end{split}
\end{equation}

Consider that there are $N_d$ datasets $D_{multi} = \{D_1, D_2, ..., D_{N_d}\}$, and thus we can obtain a set of dataset-specific language embedding $ E_{multi}=\{E_1, E_2, ..., E_{N_d}\}$ accordingly. $E_{multi}$ can be considered as an automatic ``label mapper'' by its very nature: all semantic category names of different datasets are mapped into a unified language space, where categories of similar meaning will be close while categories specific to each dataset will be represented by dedicated embeddings.

With this great property, we further convert the original index-based classification branch (dataset-specific) used in SparseRCNN into vision-language alignment based classification branch (shared). Accordingly, we adopt a region-word alignment loss rather than the traditional cross-entropy loss for optimization.
In particular, we first use a BERT model to take dataset embedding $E$, and estimate the inner relationships among different categories in each of the dataset-specific language embedding and obtain an enhanced embedding $E'$. Then we project $E'$ and object features $x^{box*}$ into a same vision-language space via two MLP branches $FC_E$ and $FC_V$ respectively. Following~\cite{clip, glip}, the alignment scores $S$ can be calculated by the dot product between $E'$ and $x^{obj}$ to reflects the similarities between visual objects and semantic categories:
\begin{align}
    &E' = \texttt{BERT}(E) \\
    &S = \texttt{Sigmoid}(FC_E(E')\cdot FC_V({x^{obj}}^*))
\end{align}
Since we treat the classification as vision-language alignment, the classification target is converted into sub-word grounding~\cite{glip} instead of one-hot label~\cite{fasterrcnn, sparse}. For each prediction and ground-truth pair, we mark the sub-word belonging to the target category as a positive pair, i.e., $\hat{T}_{i,j}=1$, otherwise $\hat{T}_{i,j}=0$.
Finally, the alignment scores can be optimized with the vision-language alignment loss $S$, same as the binary cross-entropy loss in traditional detector:
\begin{align}
    \mathcal{L}_{align}(S, \hat{T}) &= \frac{1}{N}\sum\limits_{i=1}^{N}\sum\limits_{j=1}^{L}{S_{i,j}\cdot \hat{T}_{i,j} + (1- S_{i,j})\cdot (1-\hat{T}_{i,j})} \\
    \hat{T}_{i,j} &= \left\{
    \begin{aligned}
         1 &, & \text{if} \ \text{token}_{i,j}  \, \text{belongs to} \ T_i , \\
         0 &, & \text{otherwise}
    \end{aligned}
    \right.
\end{align}
where $N$ is the number of queries, and $L$ is the length of the dataset embedding $E$.

\subsection{Dataset-aware Query}

As described above, to fully unleash the power of large amount of data in different datasets, our method proposes to use the dataset-specific language embedding to adapt the object queries so that the model can learn to adapt its behavior for each dataset. In detail, given a dataset $D$ and its language embedding $E$, we can achieve the query adaptation by simply performing a cross-attention between the learnable queries $Q$ and $E$. Then the following multi-head-self-attention is used to enhance the adapted queries.
\begin{align}
    Q_D = \texttt{Cross-Attn}(Q, E),  \quad
    {Q_D}^* = \texttt{MHSA}(Q)
\end{align}
where $Q_D$ is the adapted query for dataset $D$, and ${Q_D}^*$ is the enhanced query. Conceptually, through cross-attention, we encode the dataset-specific language embedding into the adapted queries $Q_D$, which is used as object query as Fig~\ref{fig:arch} and makes our detector dataset-aware.

Following \cite{sparse}, enhanced queries ${Q_D}^*$ and object features ${x^{obj}}$ are interacted using dynamic convolutions. In contrast to \cite{sparse}, we use enhanced queries to generate  kernels for dynamic convolutions with a few linear layers in the dynamic instance interactive head.
\begin{equation}
    \begin{aligned}
    {x^{obj}}^* &= 
    \texttt{DyConv}(x^{obj}, {Q_D}^*) \\
    &= \texttt{Conv}({K_2}({Q_D}^*), \texttt{Conv}({K_1}({Q_D}^*), x^{obj})), \\
    \quad K_i({Q_D}^*) &= \texttt{Linear}_i({Q_D}^*), i\in\{1,2\}    
    \end{aligned}
\end{equation}
where ${x^{obj}}^*$ is the interacted object feature between enhanced adapted query and object feature; $K_i(Q^*) \in \mathbb{R}^{k\times k\times c_{in}\times c_{out}}$ is the dynamic kernel generated by the enhanced adapted query ${Q_D}^*$.

To further address the challenge of position variance, besides the head, we also use a lightweight query-based RPN with one layer of convolution and the adapted queries to generate the dataset-specific convolution kernels for the RPN dynamically. Then we use the proposals of the highest $N$ scores generated by the query-based RPN as initial proposals boxes for the dynamic instance interactive head. 
The query-based RPN and decoder in our detector are both optimized in an end-to-end manner similar to \cite{sparse,dydetr,deform-detr}.

\begin{table*}[h!] 
    \centering
    \small
    \begin{tabular}{l|ccc|ccc}
    \toprule
    & \multicolumn{3}{c|}{Separate Training} & \multicolumn{3}{c}{Joint Training} \\
    & COCO & O365 & VG & COCO & O365 & VG  \\
    \midrule
    Sparse-RCNN~\cite{sparse} & 40.6 & 16.1 & 4.9 & -- & -- & -- \\
    Sparse-RCNN~\cite{sparse} + Simple merge~\cite{lambert2020mseg} & 40.6 & 16.1 & 4.9 & 40.4\small\Drop{0.2} & 17.4\small\Raise{1.3} & 4.3\small\Drop{0.6} \\
    Ours & 43.0 & 22.2 & 4.8 & 45.3\small\Raise{2.3} & 23.2\small\Raise{1.0} & 5.7\small\Raise{0.9} \\
    \bottomrule
    \end{tabular}
    \caption{\textbf{Compared with baseline methods under single and multiple dataset training.}}

\label{tab:baseline}
\end{table*}

\begin{table}[h!] 
    \centering
    \resizebox{\linewidth}{!}{
    \begin{tabular}{l|ccc|ccc}
    \toprule
    & \multicolumn{3}{c|}{Separate Training} & \multicolumn{3}{c}{Joint Training} \\
    & COCO & O365 & VG & COCO & O365 & VG  \\
    \midrule
    R-50~\cite{resnet} & 43.0 & 22.2 & 4.8 & 45.3\small\Raise{2.3} & 23.2\small\Raise{1.0} & 5.7\small\Raise{0.9} \ \\
    Swin-T~\cite{swin} & 46.3 & 25.1 & 5.6 & 49.2\small\Raise{2.9}  & 27.2\small\Raise{2.1}  & 6.4\small\Raise{0.8}  \\
    RX-101-DCN~\cite{resnext} & 49.2 & 28.3 & 6.1 & 52.0\small\Raise{2.8}  & 29.9\small\Raise{1.6}  & 7.4\small\Raise{1.3}  \\
    Swin-L~\cite{swin} & 52.5 & 35.1 & 8.0 & 55.4\small\Raise{2.9}  & 36.1\small\Raise{1.0} & 8.9\small\Raise{0.9}  \\
    \bottomrule
    \end{tabular}}
    \caption{\textbf{Comparison with different backbones under separate and joint training.}}
\label{tab:ablation-backbone}
\end{table}

\begin{table}[h!] 
    \centering
    \resizebox{\linewidth}{!}{
    \begin{tabular}{l|ccc|ccc}
    \toprule
    \multirow{2}{*}{Method} & \multicolumn{3}{c|}{Separate Training} & \multicolumn{3}{c}{Joint Training} \\
    & COCO & O365 & VG & COCO & O365 & VG  \\
    \midrule
    Instance Sampling & 15.5  & 2.3 & 0.1 & 14.5 & 2.2 & 0.1  \\
    Global Sampling & 34.5 & 19.2  & 4.8  & 39.8 & 20.2 & 5.0 \\
    Dataset-aware Sampling & 43.0 & 22.2 & 4.8 & 45.3 & 23.2 & 5.7 \\
    \bottomrule
    \end{tabular}}
    \caption{\textbf{Effectiveness of query adaption under separate and joint training.} All models are trained with the R-50 backbone.}
\label{tab:adaptation}
\end{table}

\begin{table}[h!] 
    \centering
    \resizebox{\linewidth}{!}{
    \begin{tabular}{l|ccc|ccc}
    \toprule
    \multirow{2}{*}{Query num}& \multicolumn{3}{c|}{Separate Training} & \multicolumn{3}{c}{Joint Training} \\
    & COCO & Object365 & VG & COCO & Object365 & VG  \\
    \midrule
    100  & 41.5 & 20.2 & 3.8 & 42.3 & 20.3 & 5.0 \\
    300  & 43.0 & 22.2 & 4.8 & 45.3 & 23.2 & 5.7\\
    \bottomrule
    \end{tabular}}
    \caption{\textbf{Ablation study on the number of adapted query under separate and joint training.} All models are trained with the R-50 backbone under 1$\times$ schedule.}
\label{tab:query}
\end{table}

\begin{table}[h!]
\resizebox{\linewidth}{!}{
    \begin{tabular}{l|cc | ccc }
    \toprule
    Training set & COCO & VOC & Kitti & Clipart & WaterColor\\
    \midrule
    O365 & 34.6 & 45.3 & 10.2 & 15.7 & 16.9 \\
    VG & 22.6 & 32.0 & 12.9 & 12.7 & 17.1\\
    \rowcolor{bkr} O365 + VG & \textbf{36.9} & \textbf{49.4} & \textbf{16.0} & \textbf{18.5} & \textbf{17.7}\\
    \bottomrule
    \end{tabular}}
    \caption{\textbf{Generalization capability of cross dataset evaluation.} We perform OOD evaluation to verify the generalization capability. The jointly trained model are marked in the gray rows.}
    \label{tab:cross_eval}
\end{table}

\begin{table}[h]
    \centering
    \resizebox{\linewidth}{!}{
    \begin{tabular}{ll ll lll}
        \toprule
        \multirow{2}{*}{method} && Separate && \multicolumn{3}{c}{Joint}\\
        \cmidrule{3-3} \cmidrule{5-7}
        && COCO && COCO & O365 & VG  \\
        \cmidrule{1-1} \cmidrule{3-3} \cmidrule{5-7}
        Sparse-RCNN~\cite{sparse} + Merge~\cite{lambert2020mseg} && 40.6 &&  40.4 & 17.4 & 4.3 \\
        \midrule
        + Cat-aligned Embed && 40.1\Drop{0.5} &&  43.1\Raise{2.7} & 21.3\Raise{3.9} & 5.2\Raise{0.9} \\
        + D-Q-adapt Decoder && 41.1\Raise{1.0} &&  44.0\Raise{0.9} &  21.9\Raise{0.6} & 5.4\Raise{0.2} \\
        + D-Q-adapt RPN && 43.0\Raise{1.9} &&  45.6\Raise{1.6} & 23.5\Raise{1.6} & 5.7\Raise{0.3} \\
        \bottomrule
    \end{tabular}}
    \caption{ \textbf{Baseline evolution under separate and joint training.} }
    \vspace{-4mm}
    \label{tab:component}
\end{table}

\begin{table*}[!h]  
    \centering
    \begin{tabular}{l|c|ccc|ccc}
    \toprule
     Method & ~Backbone~ & AP & AP$_{50}$ & AP$_{75}$ & AP$_{S}$ & AP$_{M}$ & AP$_{L}$ \\
    \midrule
    Deformable-DETR~\cite{borderdet} & R-50  & 37.2 & 55.5 & 40.5 & 21.1& 40.7 & 50.5\\
    Dynamic-DETR~\cite{dydetr} & R-50 & 42.9 & 61.0 & 46.3 & 24.6 & 44.9 & 54.4\\
    Sparse-RCNN~\cite{sparse} & R-50 & 39.5 & 57.7 & 42.8 & 21.8 & 42.3 & 54.4\\
    \midrule
    Ours~(separate training)& R-50 & 43.0 & 60.8 & 47.1 & 27.5 & 45.4 & 55.6 \\
    Ours~(joint training) & R-50 & 45.3 & 63.5 & 49.7 & 30.1 & 48.3& 57.3  \\
    \bottomrule
    \end{tabular}
    \caption{\textbf{Comparison with SoTA query-based models} under standard 1$\times$ setup using the ResNet-50 backbone on COCO val.}
    \label{tab:detector}
\end{table*}

\begin{table*}[!h]  
    \centering
    \resizebox{\linewidth}{!}{
    \begin{tabular}{l|ccccccccccc |c}
    \toprule
     Method & Clipart & COCO & Comic & Deeplesion & Dota & Kitchen & KITTI & LISA & VOC & WaterColor & Widerface  & Avg \\
     \midrule
     Single~\cite{wang2019towards} & 32.1 & 47.3 & 45.8 & 51.3 & 57.5 & 87.7 & 64.3 & 88.3 & 78.5 & 52.4 & 48.9 & 59.4 \\
     UniDA~\cite{wang2019towards} & 55.8 & 47.0 & 53.4 & 53.4 & 56.3 & 90.0 & 68.0 & 87.6 & 82.4 & 60.6 & 51.3 & 64.2 \\
    \midrule
    Ours & 63.5 & 56.8 & 49.7 & 62.5 & 65.2 & 94.6 & 89.3 & 97.0 & 81.3 & 57.1 & 63.7 & 71.0 \\
    \bottomrule
    \end{tabular}}
    \caption{\textbf{Ablation on UODB}, where Avg means the average score of 11 datasets. Single means the baseline trianed on each single dataset; we report the UniDA~\cite{wang2019towards} of best Avg score in the original paper. Ours means our  method jointly trained on UODB. All models take ResNet-50 as backbone. }
    \label{tab:uodb}
\end{table*}

\begin{figure*}[h!]
	\begin{center}
		\includegraphics[width=0.85\linewidth]{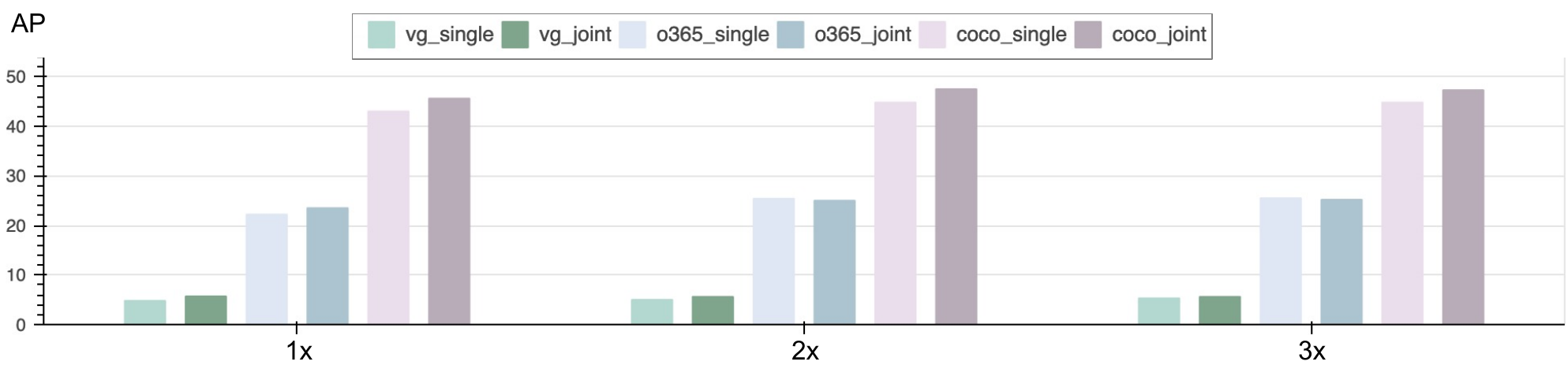}
	\end{center}
    \vspace{-0.5cm}
	\caption{Comparison with different training length under single and multiple dataset training.}
	\label{fig:tarining-length}
\end{figure*}

\begin{table*}[] 
    \centering
    \vspace{-0.2cm}
    \setlength{\tabcolsep}{3.4mm}
    \small
    \begin{tabular}{l|cc|c|c|cc}
    \toprule
    \multirow{2}{*}{Method} & \multirow{2}{*}{Training Data} & \multirow{2}{*}{Data Size} & COCO & O365 & \multicolumn{2}{c}{VG}  \\
    & & & AP & AP & AP & AP50 \\
    \midrule
    HTC++~\cite{swin} & COCO+O365 & 0.7M & 57.7 & -- & -- & -- \\
    VinVL~\cite{vinvl} & COCO+O365+VG+OID & 2.6M & -- & -- & -- & 13.7 \\
    UniDet~\cite{zhou2021simple} & COCO+O365+OID+Mapillary & 2.5M & 52.9 & 33.7 & -- & --\\
    \midrule
    Ours & COCO+O365+VG & 0.8M & 57.0 & 37.2 & 8.9 & 14.4 \\
    \bottomrule
    \end{tabular}
    \vspace{-2mm}
    \caption{Compared to SoTA on all three datasets. ``-'' indicates the numbers are not available for us.}
    \label{tab:sota}
\end{table*}

\section{Experiment}
\subsection{Implementation Detail} \label{sec:exp-detail}
We mainly evaluate our model on three popular object detection benchmarks: COCO~\cite{coco}, Object365~\cite{o365} and Visual-Genome~\cite{vg} following the common practice. COCO dataset contains 118,000 images collected from web images on 80 common classes. Object365 dataset contains around 740,000 images with 365 classes. Each image is densely annotated by human labelers to ensure quality. Visual-Genome dataset contains around 108,077 images with about 1600 classes. The images are very diverse and often contain complex categories with a long tail distribution. Since boxes are generated by algorithms, they are more noise compared to COCO and Object365. Due to different category numbers, image diversity, and annotation density, these three datasets provide a good test bed on the performance of joint training. 

We implement our method in PyTorch and train our models using V100 GPUs. We use AdamW~\cite{adamw} optimizer and step down the learning rate by a rate of 0.1 at 78\% and 93\% of epochs. For ablation studies, we use standard ImageNet-1k pre-trained ResNet-50~\cite{resnet} as the backbone with $5e^{-5}$ learning rate and $1e^{-4}$ weight decay and train it with standard 1$\times$ schedule. We demonstrate the effectiveness of each component and also compare it with other methods under this standard setup. 
For multi-dataset joint training, our detector can treat any datasets in a unified way. We only need to sample a batch from all datasets, and sample their corresponding categories to build the dataset specific language embedding. We apply dataset re-sampling to make sure that each dataset was trained at only 1$\times$ schedule. We also apply category re-balancing to handle the long-tailed distributions of Object365. Besides, we also evaluate our method on a variety of backbones to demonstrate robustness. For experiments comparing with state-of-the-art methods, we train our method with Swin Large~\cite{swin} backbone at the 2x schedule with multi-scale training. We set the learning rate and weight decay the same as ResNet-50. During evaluation, we evaluate our best model with multi-scale testing to compare with state-of-the-art methods reported without using test time augmentation. There is no other augmentation or optimization tricks used during training.

\subsection{Effectiveness of Multi-dataset Training}
For the baseline, we build a Sparse R-CNN model with a ResNet-50 and train on three single-dataset separately. As for baseline for multi-dataset joint training, we follow~\cite{wang2019towards}, which simply concatenates all datasets and maps their taxonomy together. Then we can train it with a bigger label mapping layer with a common loss. For our proposed method, we evaluate both joint and separate training ways. 

As shown in Table~\ref{tab:baseline}, under standard 1$\times$ schedule, the jointly trained baseline model is observed a clear performance drop on COCO and VG compared to models trained separately. It demonstrated that simply mapping different labels together doesn't work well due to the domain gaps and taxonomy differences between the datasets. On the contrary, our query-adapted joint training obtains higher performance on each dataset, especially offering a significant gain at 2.3, 1.0, and 0.9 mAP gain on COCO, Object365, and Visual-Genome, respectively. The experiment well demonstrated that our single detector can leverage benefits from multiple datasets training simultaneously. 

We also train our method with multiple backbones, such as ResNet-50, ResNext-101, Swin-Tiny, Swin-Large to verify its robustness. As shown in Table \ref{tab:ablation-backbone}, it is clear to see that our method obtains significant performance gains on all datasets when jointly trained compared to separate trained. 

Furthermore, we also evaluate our method's multi-dataset training stability under different training lengths. As shown in Fig~\ref{fig:tarining-length}, the performance gain preserves even when prolonging the training beyond full convergence. These experiments well demonstrate the effectiveness and robustness of our method under multiple datasets joint training.

\subsection{Effectiveness of Query Adaptation}
We then evaluate the effectiveness of our proposed query adaption method. To compare, we design two baselines: ``Instance Sampling'' means only considering categories belonging to each training image. This limits the usage of category embedding to image level, where categories from different images are not interacted with each other; ``Global Sampling'' means we consider the categories from all datasets. This allows all categories in different datasets to interact with each other. As shown in Table \ref{tab:adaptation}, query adaption is the key to effective multi-dataset training. With ``Global Sampling'', separate and joint training drops significantly due to the negative interfering between different datasets. Finally, our dataset-aware query can support ``Dataset-aware Sampling'', which samples category names according to the source dataset of each training image specifically to improve performance.

In addition, we also conduct an ablation on utilizing different numbers of adapted queries. As shown in Table~\ref{tab:query}, it is obvious that more adapted queries further enlarge the performance gains from joint training compared to separate training. To further verify the generalization capability of our adapted queries, we conduct the cross dataset evaluation by evaluating on five datasets out of the training set as shown in Table~\ref{tab:cross_eval}. Our models, especially the joint-training model, obtain non-trivial performance.
Overall, above ablation studies well justify the effectiveness of our query adaptation.  

\subsection{Effectiveness of Detector}
We also evaluate our query adapted detector separately to compare it with popular detectors. To conduct a fair comparison and ensure reproduction, we first train our method with a ResNet-50 on COCO data only using a standard 1$\times$ schedule without augmentation and multi-dataset training. We compare our result with state-of-the-art query-based methods reported in a similar setting, such as~\cite{deform-detr,sparse,dydetr}. Shown in Table ~\ref{tab:detector}, our method achieves state-of-the-art performance at 43.0 AP. In addition, our method really shines when trained under multi-dataset joint training and outperforms previous methods by 
clear margins.

\subsection{Effectiveness of Each Component}
To further clarify the baseline evolution from Sparse-RCNN to our model, we study the effectiveness of each component under separate and joint training.
The two main differences can be summarized as follows: 1) \emph{Category-aligned Embedding} replaces the traditional classifier through visual-word alignment and 2) \emph{Dataset-aware Query} is incorporated for both the decoder and the extra RPN. As shown in Tab~\ref{tab:component} Such a combination offers a 2.4 AP improvement on COCO under separate training and best boost the detector under joint training.

\subsection{Effectiveness on Extreme Varied Datasets}
We further evaluate on UODB~\cite{wang2019towards}, a combination of COCO~\cite{coco}, KITTI~\cite{kitti}, WiderFace~\cite{widerface}, VOC~\cite{voc}, LISA~\cite{lisa}, DOTA~\cite{dota}, Watercolor~\cite{pda}, Clipart~\cite{pda}, Comic~\cite{pda}, Kitchen~\cite{kitchen}, and DeepLesion~\cite{deeplesion}. These datasets cover wide variations in category, camera view, image style, etc. Following UODB~\cite{wang2019towards}, we train our model with ResNet-50 as backbone and report the $AP_{50}$ for each dataset separately. As shown in Table~\ref{tab:uodb}, our Detection Hub can outperform UDA~\cite{wang2019towards} by a large margin on avg score. It provides a convincing results, highlighting the effectiveness on the combinations of extremely varied datasets.

\begin{figure*}[!htb]
	\begin{center}
		\includegraphics[width=0.82\linewidth]{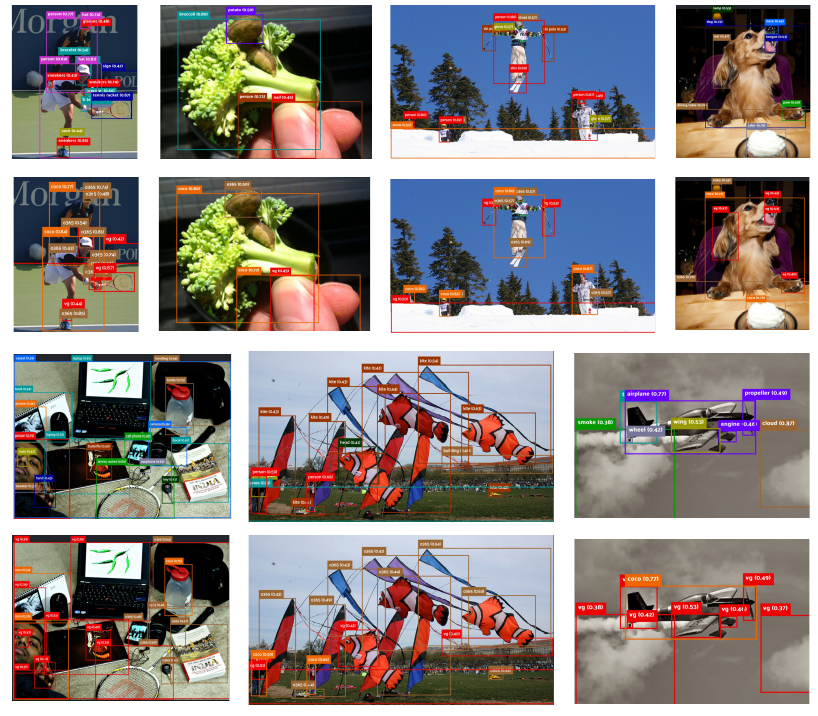}
	\end{center}
	    \vspace{-0.1in}
	\caption{Visualization of different predictions results from our method. The first odd rows show the predicted category names. The even rows show which dataset contributes to our final prediction. }
	\label{fig:viz}
    \vspace{-6mm}
\end{figure*}

\subsection{Comparison with SoTA}
Finally, we compare our method with state-of-the-art object detectors using a large backbone. We report our best performance with Swin Large jointly trained on COCO, Object365, and Visual-Genome datasets. Unlike previous methods that require separate pre-training and fine-tuning steps, our method directly reports the performance after joint training without domain-specific fine-tuning. As shown in Table~\ref{tab:sota}, we compare with recent methods that leverage multiple datasets. Compared to previous best results, our method archives new state-of-the-art results on both Object365 and Visual-Genome. On COCO, our method is competitive with recent work HTC++~\cite{swin}. However, HTC++ is first pre-trained on Object365 and COCO, then finetuned on COCO for a better result. Besides, unlike HTC++ combining the segmentation task and detection task, our method achieves this performance without the help of instance segmentation information. Meanwhile, our method demonstrates significant improvements on both COCO and Object365 over concurrent work~\cite{zhou2021simple}. 
In conclusion, thanks to our effective query adaption under multi-dataset joint training, our method can take advantage of different data to further advance the state-of-the-art results.

\subsection{Visualization}
Finally, we visualize the predictions in Fig~\ref{fig:viz}, where the odd rows show the predicted category names and the even show which dataset contributes to the prediction. 
Our predictions effectively combine the taxonomies from multiple datasets and yield better predictions. 
This well justifies our motivation for proposing a universal object detector.

\section{Conclusion}
Unifying multiple object detection datasets with a single object detector is a non-trivial problem. We propose \emph{``Detection Hub"} to address the inherent taxonomy differences and annotation inconsistency challenges. It can enjoy the synergy of multiple datasets and achieve substantial improvements over the independently trained detector baseline. 
Meanwhile, we also find that the performance on the dataset with a large vocabulary may be constrained by the maximum length of language embedding. 
In the future, we will try to expand our work to open-world object detection by combining more datasets from different domains to cover a wide variety of concepts. 

\vspace{0.01in}
\noindent\textbf{Acknowledgement} This project was supported by National Key R\&D Program of China (No. 2021ZD0112805).

\newpage
{\small
\bibliographystyle{abbrv}
\bibliography{reference}
}
\end{document}